# Quantitative Characterization of Retinal Features in Translated OCTA

**Rashadul Hasan Badhon[1], Atalie Carina Thompson[2], Jennifer I. Lim[3], Theodore Leng[4], Minhaj Nur Alam[1, *]**

[1] *Department of Electrical and Computer Engineering, University of North Carolina at Charlotte, Charlotte, NC, United States*

[2] *Department of Surgical Ophthalmology, Atrium-Health Wake Forest Baptist, Winston-Salem, NC, United States*

[3] *Department of Ophthalmology and Visual Science, University of Illinois at Chicago, Chicago, IL, United States*

[4] *Department of Ophthalmology, Stanford University School of Medicine, Stanford, CA, United States*

[*]Correspondence:

Minhaj Nur Alam, PhD

9201 University City Boulevard, Charlotte, NC 28223

minhaj.alam@charlotte.edu

**Funding:** Supported by NEI R15EY035804 (MNA) and UNC Charlotte Faculty Research Grant (MNA).

**Commercial Relationship Disclosure:** There is no commercial relationship from any authors related to this research work.

# Abstract

**Purpose:** This study explores the feasibility of using generative machine learning (ML) to translate Optical Coherence Tomography (OCT) images into Optical Coherence Tomography Angiography (OCTA) images, potentially bypassing the need for specialized OCTA hardware.

**Methods:** The method involved implementing a generative adversarial network framework that includes a 2D vascular segmentation model and a 2D OCTA image translation model. The study utilizes a public dataset of 500 patients, divided into subsets based on resolution and disease status, to validate the quality of TR-OCTA images. The validation employs several quality and quantitative metrics to compare the translated images with ground truth OCTAs (GT-OCTA). We then quantitatively characterize vascular features generated in TR-OCTAs with GT-OCTAs to assess the feasibility of using TR-OCTA for objective disease diagnosis.

**Result:** TR-OCTAs showed high image quality in both 3 and 6 mm datasets (high-resolution, moderate structural similarity and contrast quality compared to GT-OCTAs). There were slight discrepancies in vascular metrics, especially in diseased patients. Blood vessel features like tortuosity and vessel perimeter index showed a better trend compared to density features which are affected by local vascular distortions.

**Conclusion:** This study presents a promising solution to the limitations of OCTA adoption in clinical practice by using vascular features from TR-OCTA for disease detection.

**Translation relevance:** This study has the potential to significantly enhance the diagnostic process for retinal diseases by making detailed vascular imaging more widely available and reducing dependency on costly OCTA equipment.

Optical Coherence Tomography (OCT) is a cutting-edge medical imaging technology that has revolutionized our ability to observe and comprehend the complex structures of the human body. It is non-invasive and capable of providing highly detailed in-depth retinal pathologies. It generates high-resolution cross-sectional images of tissues using low-coherence light, therefore has been widely adopted in ophthalmic clinical care.[1] As a result, OCT has been demonstrated for early identification and monitoring of various retinal illnesses including diabetic retinopathy (DR), age-related macular degeneration (AMD) and glaucoma that cannot be obtained by any other non-invasive diagnostic technique.[2–8]

The rapid development of OCT, growing interest in this field, and its increasing impact in clinical medicine has contributed to its widespread availability. However, due to its non-dynamic imaging technology, OCT cannot visualize blood flow information such as blood vessel caliber or density and remains only limited to capturing structural information.[2,9] As a result of this information gap, OCT angiography (OCTA) was developed which can produce volumetric data from choroidal and retinal layers and provide both structural and blood flow information.[10,11] OCTA provides a high-resolution image of the retinal vasculature at the capillary level, allowing for reliable detection of microvascular anomalies in diabetic eyes and vascular occlusions. It helps to quantify vascular impairment based on the severity of retinal diseases. In recent years, OCTA has been demonstrated to identify, detect, and predict DR,[12–19] AMD,[20–22] Glaucoma[23] and several other retinal diseases.[24–31] Despite the advantages, widespread deployment of OCTA has been limited due to the high device cost.[32,33] The additional requirements of hardware and software for an OCTA device pose a financial burden for clinics as well as patients, therefore, there are only a limited number of hospitals and retinal clinics that use OCTA on a daily basis, for routine ophthalmic check-ups. Another limitation of OCTA is the process of generating an OCTA scan, which takes longer time and involves repetitive scanning of the retina making the data acquisition harder due to involuntary eye movements and motion artifacts, reducing the quality of OCTA images.[33] Due to the limitation of OCTA data, most studies involving OCTA based imaging biomarkers and involving the use of artificial intelligence (AI) are difficult to validate extensively for future clinical deployment.

From literature, a potential solution to this problem can be the utilization of AI and machine learning (ML) to produce OCTA images from the already available OCT data which has been showing promising outcomes.[34–38] Incorporating ML for OCTA translation from OCT offers significant advances in ophthalmic diagnostics by increasing angiographic and functional information in existing OCT data. This transition harnesses ML's capability to autonomously analyse OCT scans and generate detailed vascular images, traditionally obtained through OCTA, aligned with OCT information. By doing so, it substantially lowers the barriers to accessing high-resolution vascular imaging, which is crucial for diagnosing and monitoring retinal diseases and provides a robust detection system. Furthermore, ML dependent approaches alleviate some of OCTA's limitations, including its high cost, susceptibility to artifacts from patient movement and the extensive time required for image acquisition.

Different studies have been reported[39–41] attempting to leverage ML algorithms for generative-adversarial learning, typically utilizing a UNet for image translation in recent years. However the quality of the translated OCTA (TR-OCTA) is usually sub-optimal and the retinal vascular areas are not refined enough. The first application of this approach was reported by C. S. Lee et al., 2019[34] to train an algorithm to generate retinal flow maps from OCT images avoiding the needs for labelling but it was limited to capture higher density of deep capillary networks. According to some recent studies,[35–37] incorporating textual information or surrounding pixels, it is possible to improve the OCTA image quality. In this paper, we adopt and implement a generative-adversarial learning framework-based algorithm demonstrated by Li et. al [36] for translating OCT data into OCTA. of vascular regions of translated OCTA images. The focus of this study is to demonstrate the feasibility of using such TR-OCTA image generated vascular features (Blood Vessel Density (BVD), Blood Vessel Caliber (BVC), Blood Vessel Tortuosity (BVT), Vessel Perimeter Index (VPI)) for disease detection. We compare these OCTA features with ground truth (GT) – OCTAs. The quality of the GT-OCTAs were compared with features such as Structural Similarity Index Measure (SSIM), Fréchet Inception Distance (FID) and patch-based contrast quality index (PCQI). From our observation and statistical analysis, we found that overall, the SSIM values indicate a moderate level of structural similarity between TR-OCTA and GT-OCTA images, with some variability across different patient categories and resolutions however PCQI scores are quite close for both dataset and some deviation in FID scores is noticeable. It was observed that the model generally achieved a slightly better performance in depicting normal and pathological retinal features at the 3mm resolution compared to the 6mm resolution. However, across both resolutions, there were slight discrepancies in quantitative vascular metrics such as BVD, BVC and VPI, highlighting areas where the translation model could be further refined. This analysis underscores the potential of using AI-driven translation models for OCTA image analysis, while also pointing to the need for improvements to enhance the accuracy of vascular feature representation, particularly at varying resolutions.

## Methodology

The overall methodology of our feature extraction pipeline is demonstrated in Fig. 1. We first translate OCTA from our OCT data (using algorithm demonstrated by Li et. al [36]) and quantify the retinal features in both GT and TR-OCTAs for validation. Fig. 2 and Fig. 3 show the GT-OCTA and TR-OCTA images for both 3mm and 6mm datasets for diseased as well as normal patients.

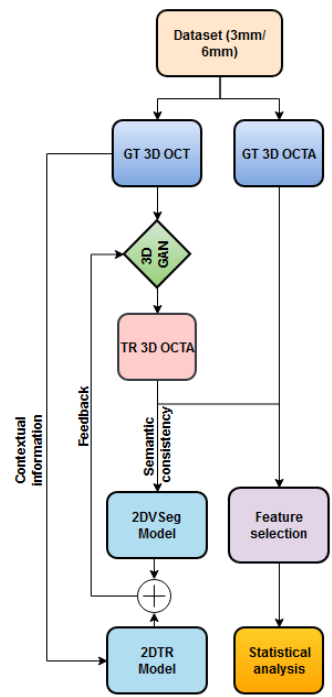

Fig. 1: Framework of OCT to OCTA translation and characterization of quantitative features.

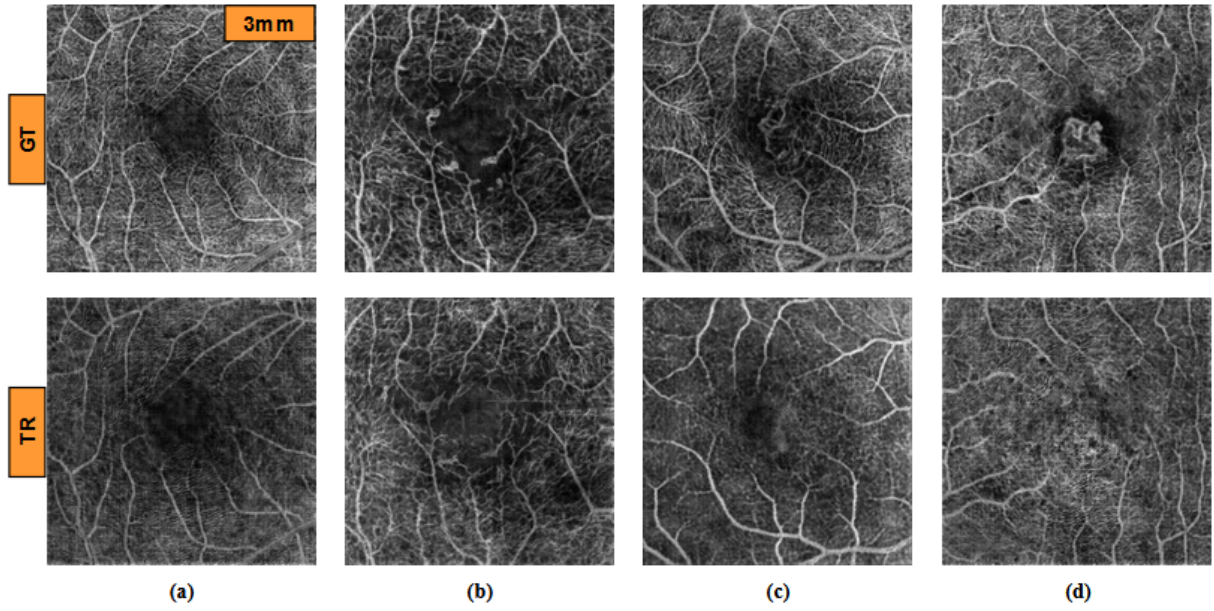

Fig. 2: GT and TR OCTA images from (a) Normal, (b) DR, (c) CNV and (d) AMD patients for 3mm dataset.

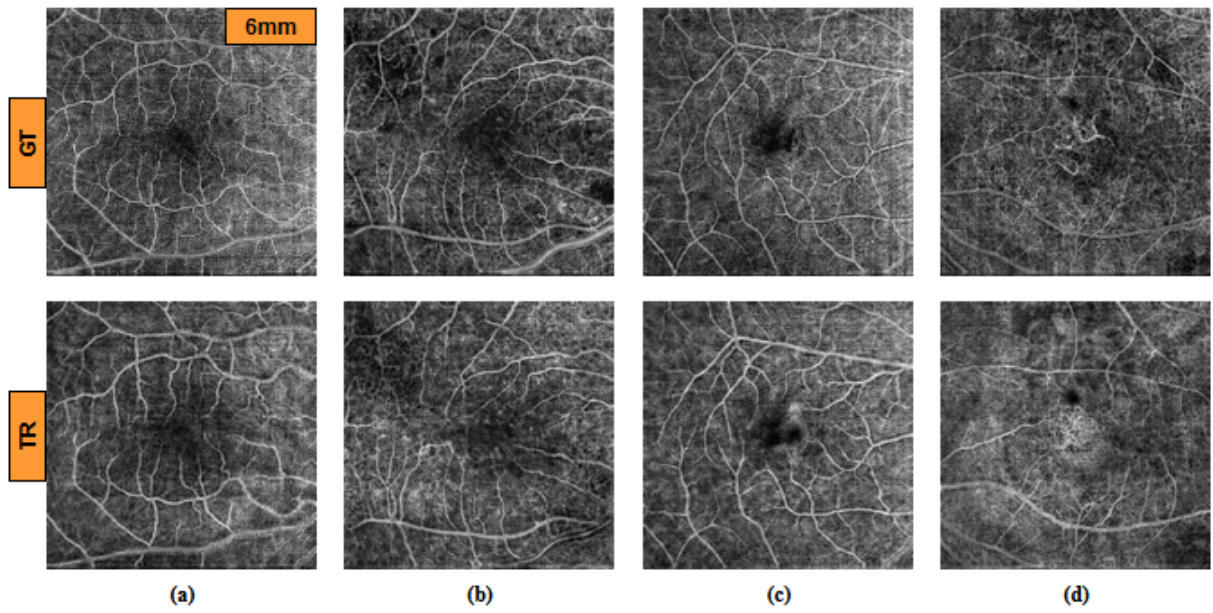

Fig. 3: GT and TR OCTA images from (a) Normal, (b) DR, (c) CNV and (d) AMD patients for 6mm dataset.

**Dataset**

We used a public dataset of 500 patients containing paired 3D OCT and OCTA volumes, OCTA-500.[42] The dataset is divided into 2 subsets according to their resolutions, 3mm and 6mm. The translation algorithm is applied separately to the two subsets for comparison. The datasets are further divided into different diseased patients and normal patients for quantitative feature comparison: SSIM, BVD, BVC, BVT and VPI. This whole dataset contains 6 AMD patients, 5 Choroidal neovascularization (CNV) patients, 29 DR and 160 Normal patients who are divided according to the diseases and compared statistically after evaluating the feature values.

The set contains paired OCT and OCTA volumes from 200 patients with a field of view (FOV) 3mm × 2mm × 3mm. Each volume has 304 slices with a size of 640px x 304px. The generated projection map is of 256px x 256px size. The whole dataset is divided into a 70-30% split: 140, 10 and 50 volumes for training, validation and test sets respectively. Similarly, this set contains paired OCT and OCTA volumes from 300 patients with FOV of 6mm × 2mm × 6mm. Each volume is of size 640px × 400px, containing 400 slices and generated projection maps are of size 256px x 256px. Similar to 3mm set: 180, 20 and 100 volumes are split as training, validation and test sets. The 6mm dataset contains 43 AMD, 11 CNV, 14 Central serous chorioretinopathy (CSC), 35 DR, 10 Retinal vein occlusion (RVO), 91 Normal and 96 other retinal pathology-affected patients for which a similar statistical evaluation is carried out and feature values are calculated.

**Translation algorithm**

We adopted and implemented the OCT to OCTA translation algorithm from Li et. al [36]. We describe the process here briefly. The process of OCTA translation from OCT images is carried out in 3 steps (Fig. 1): (a) generating 3D OCTA volumes from paired 3D OCT volumes using conditional generative adversarial network (GAN), (b) improving image quality by focusing only the vascular regions, utilizing the 2DVSeg model, thorough vascular segmentation, (c) preserving contextual information for better quality translated images through a 2D translation model (2DTR) generating 2D paired OCTA maps. The baseline architecture of the translation model is built upon pix2pix, an image translation model [40]. The aim of the model is primarily to translate OCT volumes, $X$ to its paired OCTA volume $\hat{Y}$ as closely as possible to the original clinical images, OCTA volume, $Y$.[4] The framework includes a 3D GAN where the 3D generator takes a 3D OCT volume as its input and outputs a corresponding TR-OCTA volume. a 3D discriminator is used to effectively distinguish between the original (ground-truth) OCTA volumes and the generated ones. These components are referred to as $G_{3d}$ for the generator and $D_{3d}$ for the discriminator. An adversarial loss is used to train both the generator and discriminator. Furthermore, to calculate for each pixel difference between TR-OCTA and GT-OCTA, a distance loss is considered    The framework also uses a 2D vascular segmentation model (Fig. 4) to help with the improved quality of the vascular regions by utilizing OCTA reflected vascular data. that focuses on the vascular areas during the 3D volume translation process. that focuses on the vascular areas during the 3D volume translation process.

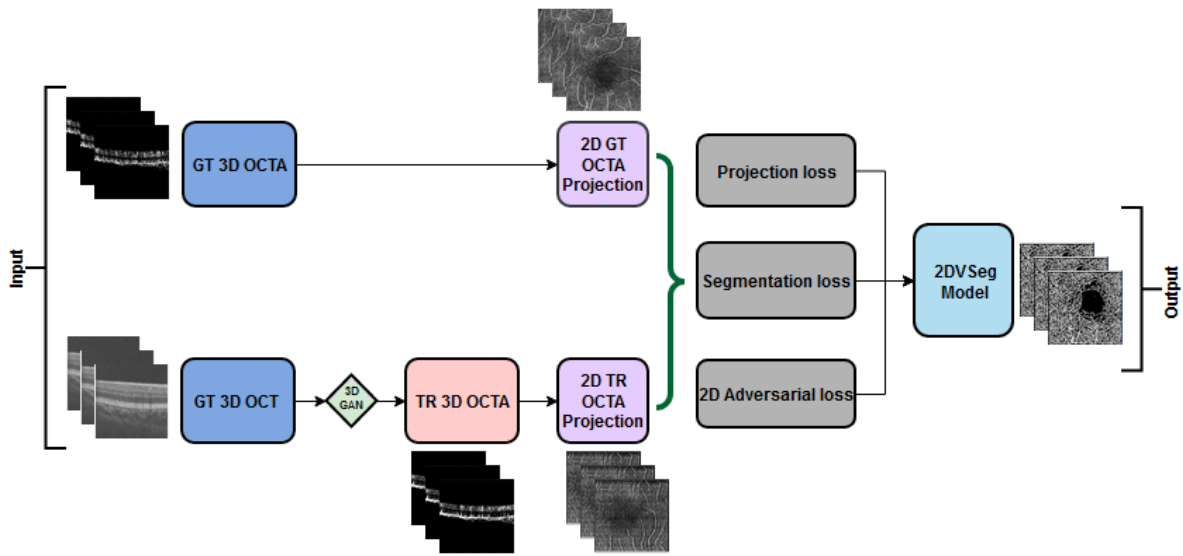

Fig. 4: 2D Vascular segmentation model

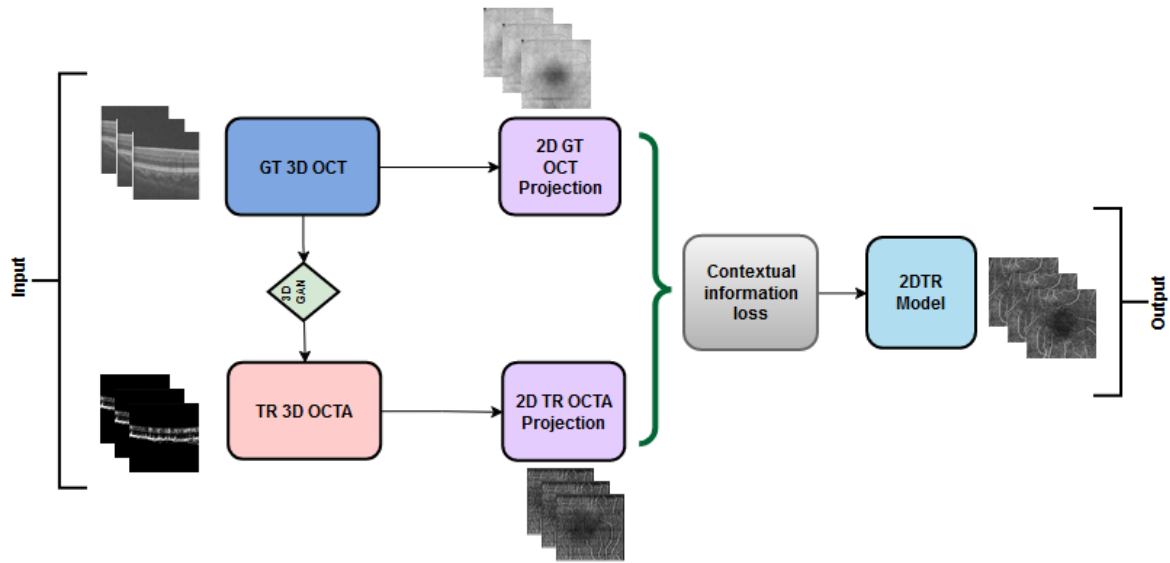

Fig. 5: 2D Translation model

This model also utilizes a 2D generative translation model (Fig. 5) to build heuristic (suboptimal) 2D OCTA projection maps from their corresponding OCT that can provide heuristic contextual information where output values are affected by the surrounding pixels resulting in outputs with additional contextual information. The TR-OCTA maps that were generated were then compared on several quantitative features to the GT projection maps for comparison: BVD, BVC, BVT and VPI. SSIM, FID and PCQI metrics were used to quantify the quality and similarity to GT OCTA maps.

**Metrics and Features**

**SSIM:** SSIM or Structural Similarity Index Measure, is a method for measuring the similarity between two images. SSIM is based on the perception of the human visual system and it considers changes in structural information, luminance and contrast. The idea is that pixels have strong inter-dependencies, especially when they are spatially close. These dependencies carry important information about the structure of the objects in the visual scene.

**BVD**: BVD or vessel area density (VD),[44] is the ratio of the blood vessels to the total area measured [26] and can be utilized for identifying early detection of retinal pathologies including DR,[45,46] AMD[47,48] etc.

$$BVD = \frac{vascular\ area}{total\ area} \quad (15)$$

**BVC:** BVC, also named as vessel diameter index[49], is calculated as the ratio of vessel area to the vessel length.[12] BVC distortion can be used to quantify retinal vascular shrinkage and is typically observed in different retinopathies such as diabetic retinopathy (DR).[50]

$$BVC = \frac{vascular\ area}{vascular\ length} \quad (16)$$

**BVT**: BVT is defined as a measure of degree of vessel distortion.[26,51] During any retinal pathologies, distorted vessel structures can affect the blood flow efficiency and can be measured as:

$$BVT = \frac{1}{n} \sum \frac{geodesic\ distance\ between\ endpoints\ for\ a\ vessel\ branch}{euclidean\ distance\ between\ endpoints\ for\ a\ vessel\ branch} \quad (17)$$

here, n = total number of vessel branches

**VPI:** VPI[51] is measured as the ratio of the contour length of the vessel boundaries or vessel perimeter to the total vessel area and has been used for detection of DR and sickle cell retinopathy (SCR) from OCTA images:

$$VPI = \frac{overall\ contour\ length\ of\ blood\ vessel\ boundaries}{total\ blood\ vessel\ area} \quad (18)$$

Statistical Analysis: We performed statistical analysis based on the selected features to quantify the TR-OCTA and measure the quality of the translation. This analysis will help us improve the accuracy and efficiency of the TR-OCTA translated from GT-OCT and GT-OCTA.

**FID & PCQI:** FID score is a metric used to evaluate the quality of images generated by models, such as those produced by GANs. It measures the similarity between two sets of images, typically between a set of generated images and a set of real images, by comparing the statistics of their features extracted by a pre-trained Inception model.[52] The FID score calculates the distance between the feature vectors of the real and generated images, using the Fréchet distance (also known as the Wasserstein-2 distance). A lower FID score indicates that the distribution of the generated images is closer to the distribution of the real images, suggesting higher quality and more realistic images.

PCQI is another metric designed to assess the quality of images by focusing on local contrast changes, which are crucial for visual perception, especially in textured regions.[53] Unlike many traditional image quality metrics that evaluate images globally, PCQI operates on small, localized patches of an image, making it particularly effective at capturing and evaluating detailed contrast differences between a reference image and a test image. PCQI calculates the quality score based on three main aspects: patch similarity, contrast distortion, and mean luminance change, within these localized regions. The final score is a weighted sum of these aspects, providing a single quality metric that reflects how perceptually close the test image is to the reference image in terms of local contrast and brightness. A higher PCQI score indicates a better match between the test and reference images, suggesting less contrast distortion and more accurate reproduction of the original image's visual quality.

## Results

TABLE 1, TABLE 2 and TABLE 3 summarize the translated OCTA quality metric analysis and quantitative OCTA feature analysis for both 3mm and 6mm datasets respectively. For the 3mm dataset, SSIM was found to be ranging from 0.29-0.60 with a mean of 0.48 and 6mm dataset showed SSIM ranging from 0.16-0.52 with a mean of 0.42. We also calculated SSIM values for comparing different patient statuses for both datasets. From 3mm: AMD patients show a slightly lower mean SSIM score of 0.4513, CNV patients exhibit an SSIM mean of 0.4754 with a narrower range, DR dataset on the other hand reveals a higher mean SSIM score of 0.4923 and finally, the Normal group shows an SSIM mean of 0.4834. Similarly, when calculated for 6mm: AMD, CNV, CSC, patients with other

retinopathies and Normal group showed a close SSIM mean within a range of 0.41-0.42 with exceptions of DR patients having slightly higher SSIM (0.43) and RVO having smaller mean of 0.36. Furthermore, TABLE 3 presents FID and PCQI scores for OCTA datasets at two different resolutions, 3mm and 6mm. FID shows a lower score (35.88) for the 3mm dataset, indicating closer resemblance to real images compared to the 6mm dataset, which has a higher FID score of 49.06. On the other hand, the PCQI scores, assessing image quality in terms of contrast and sharpness, are comparably high for both datasets, with the 3mm dataset slightly outperforming the 6mm dataset (0.99795±0.000457 vs. 0.99778±0.000539). This suggests that, despite the higher FID score, the 6mm dataset maintains a high level of contrast quality, albeit slightly lower than its 3mm counterpart.

Two-tail T-tests were carried out ($\alpha<.05$) for BVD, BVC, BVT and VPI (3mm complete dataset) but only BVD and BVC proved to be containing statistically similar values: BVD (0.48), BVC (0.45), BVT ($1.1e^{-7}$) and VPI ($1.36e^{-22}$). BVD for TR-OCTA and GT-OCTA are 212.31±29.93 vs 210.22±29.044 respectively for the whole dataset. We also calculated BVC: (22.80±0.81 vs 22.75±0.41), BVT: (1.086±0.0064 vs. 1.089±0.0061) and VPI: (26.91±5.47 vs. 31.43±2.35) for TR-OCTA and GT-OCTA respectively. Additionally, these features are calculated separately for different diseased and Normal patients. For AMD, DR and Normal patients, BVD was found to be closely aligned to the results we got for the complete dataset, compared to CNV patients (228.53±22.36; 224.22±16.47). SSIM values were measured within a range of 0.47-0.50 for diseased as well as Normal patients. BVC and BVT had similar values for all cases compared to VPI having a wider difference between TR-OCT and GT-OCT. Overall, TR-BVC, TR-VPI, TR-BVT and TR-BVD values (Fig. 6) are concentrated within a specific range and closer to the GT values for each feature respectively. BVC, VPI and BVD have some outliers, specifically for BVD, some outliers are further away from the lowest value of the BVD range.

For comparison among different diseased and normal patients, Supplemental Fig. 1 (a-d) represents the distribution of BVC, VPI, BVT and BVD feature values for AMD, CNV, DR and Normal subjects respectively. In comparison to other features (Supplemental Materials), BVD is found to contain more outliers for normal patients rather than the diseased patients which was also prominent in the feature values of the whole 3mm dataset. Similarly, for the 6mm dataset (TABLE 2), we performed T-tests ($\alpha<.05$) for BVD, BVC, VPI and BVT but only BVD was found to have statistically similar values for both TR-OCTA and GT-OCTA images: BVD (0.58), BVC ($1.35e^{-52}$), BVT (0.006), VPI ($8.26e^{-31}$). For BVD we calculated 210.80±30.45 and 212.34±37, BVC 22.80±.81 and 42.91±1.33, BVT 1.0868±.0065 and 1.088±.007, VPI 24.95±2.969 and 27.937±3.019 for TR-OCTA and GT-OCTA respectively (Fig.6). The 6mm dataset contained central serous chorioretinopathy (CSC), retinal vein occlusion (RVO) and other retinal pathologies that were absent in the 3mm dataset. In a comparative analysis among diseased and normal patients, SSIM and BVD for RVO patients showed a larger deviation compared to other cases when calculated. However, BVC, BVT and VPI were measured having closer values in all cases.

For the complete dataset, Fig. 6 shows the distribution of the feature values for TR-OCTA and GT-OCTA. Similar to the 3mm dataset, BVC, VPI and BVD have more outliers compared to BVT and the distribution is similar to the 3mm dataset. Supplemental Materials include boxplots of BVC, VPI, BVT and BVD feature values of diseased patients as well as normal patients. Supplemental Fig. 2 (a-g) represents feature values for AMD, CNV, CSC, DR, RVO, other retinal pathologies and normal patients. A similar trend of BVD feature having more outliers is noticeable for diseased as well as normal patients in comparison to other features (Appendix B) except RVO.

TABLE 1: Statistical analysis of TR-OCTA compared to GT-OCTA for 3mm dataset.

| OCTA Dataset | SSIM (Mean & range) | Dataset (no. of patients) | BVD (Mean±St.d) | BVC (Mean±St.d) | BVT (Mean±St.d) | VPI (Mean±St.d) |
|---|---|---|---|---|---|---|
| TR-OCTA | 0.4835 | Complete (200) | 212.31±29.93 | 22.80±0.81 | 1.086±0.006 | 26.91±5.47 |
| GT-OCTA | (0.29-0.60) | | 210.22±29.04 | 22.75±0.41 | 1.089±0.006 | 31.43±2.35 |
| TR-OCTA | 0.4513 | AMD (6) | 213.73±20.05 | 22.45±1.03 | 1.087±0.009 | 29.24±1.91 |
| GT-OCTA | (0.29-0.55) | | 205.46±26.45 | 22.91±0.39 | 1.09±0.003 | 29.69±1.63 |
| TR-OCTA | 0.4754 | CNV (5) | 228.53±22.36 | 22.34±1.04 | 1.087±0.003 | 26.36±3.7 |
| GT-OCTA | (0.44-0.52) | | 224.22±16.47 | 22.90±0.53 | 1.089±0.005 | 30.26±2.55 |
| TR-OCTA | 0.4923 | | 209.07±27.51 | 23.12±0.71 | 1.080±0.007 | 26.92±4.32 |

| OCTA Dataset | SSIM (Mean) | Dataset (no. of patients) | BVD (Mean±St.d) | BVC (Mean±St.d) | BVT (Mean±St.d) | VPI (Mean±St.d) |
|---|---|---|---|---|---|---|
| GT-OCTA | (0.29-0.59) | DR (29) | 210.80±34.82 | 23.14±0.42 | 1.087±0.005 | 28.25±3.55 |
| TR-OCTA | 0.4834 | NORMAL (160) | 212.34±30.86 | 22.77±0.81 | 1.086±0.006 | 26.84±5.79 |
| GT-OCTA | (0.34-.60) | | 209.86±28.39 | 22.67±0.37 | 1.089±0.006 | 32.11±1.41 |

TABLE 2: Statistical analysis of TR-OCTA compared to GT-OCTA for 6mm dataset.

| OCTA Dataset | SSIM (Mean) | Dataset (no. of patients) | BVD (Mean±St.d) | BVC (Mean±St.d) | BVT (Mean±St.d) | VPI (Mean±St.d) |
|---|---|---|---|---|---|---|
| TR-OCTA | 0.4175 | Complete (300) | 210.80±30.45 | 44.78±1.37 | 1.087±0.006 | 24.95±2.97 |
| GT-OCTA | (0.16-0.52) | | 212.34±37 | 42.91±1.33 | 1.088±0.007 | 27.94±3.02 |
| TR-OCTA | 0.4102 | AMD (43) | 210.13±32.07 | 44.28±1.28 | 1.063±0.005 | 24.56±3.32 |
| GT-OCTA | (0.30-0.50) | | 204.72±35.76 | 42.93±1.44 | 1.063±0.007 | 27.76±3.65 |
| TR-OCTA | 0.4224 | CNV (11) | 213.11±27.01 | 45.00±1.03 | 1.087±0.003 | 24.06±1.63 |
| GT-OCTA | (0.38-0.45) | | 224.63±46.59 | 42.59±1.03 | 1.089±0.007 | 27.39±3.36 |
| TR-OCTA | 0.4140 | CSC (14) | 209.66±22.43 | 45.12±0.96 | 1.088±0.0064 | 25.08±1.86 |
| GT-OCTA | (0.32-0.45) | | 215.35±45.51 | 43.08±0.98 | 1.088±0.0063 | 28.59±2.24 |
| TR-OCTA | 0.4329 | DR (35) | 215.09±28.04 | 45.12±1.30 | 1.086±0.0065 | 26.2±2.95 |
| GT-OCTA | (0.35-0.52) | | 210.66±37.54 | 43.50±1.18 | 1.087±0.0068 | 28.68±3.2 |
| TR-OCTA | 0.3664 | RVO (10) | 228.13±60.63 | 44.13±1.15 | 1.089±0.0079 | 24.82±1.76 |
| GT-OCTA | (0.26-0.43) | | 239.79±26.76 | 43.09±0.91 | 1.087±0.009 | 27.55±2.89 |
| TR-OCTA | 0.4169 | Others (96) | 207.14±30.19 | 44.88±1.34 | 1.086±0.0062 | 25.19±3.06 |
| GT-OCTA | (0.16-0.51) | | 213.22±32.52 | 43.06±1.19 | 1.088±0.0073 | 27.74±2.61 |

| | | | 211.34±27.64 | 44.75±1.50 | 1.087±0.0072 | 24.5±2.97 |
| TR-OCTA | 0.4212 | Normal | | | | |
| GT-OCTA | (0.25-0.49) | (91) | 210.71±39.42 | 42.53±1.51 | 1.089±0.0073 | 27.95±3.13 |

TABLE 3: FID and PCQI scores for the complete datasets of 3mm and 6mm

| OCTA Dataset | FID | PCQI (Mean±St.d) |
| --- | --- | --- |
| 3mm | 35.88 | 0.99795± 0.000457 |
| 6mm | 49.06 | 0.99778± 0.000539 |

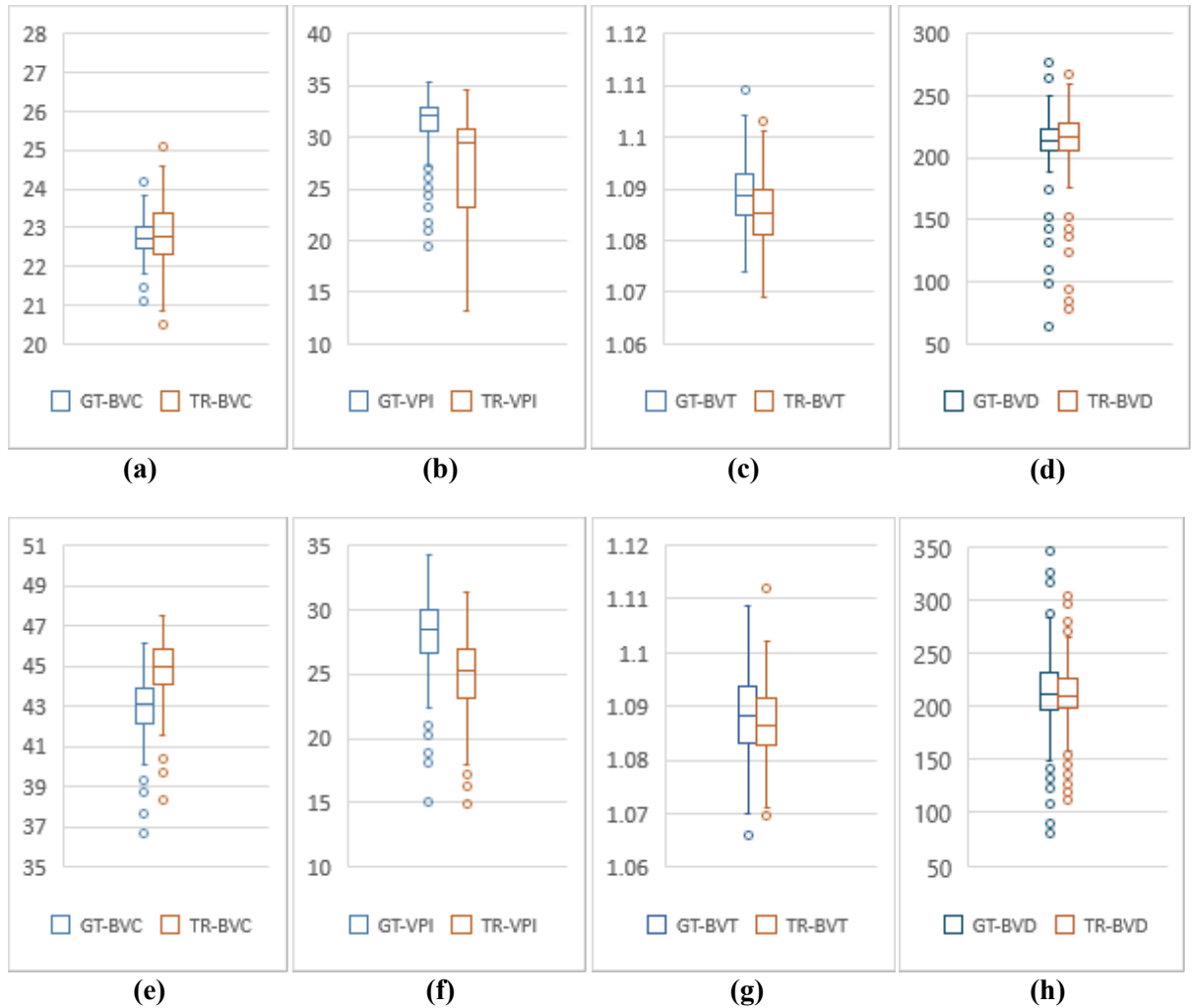

**Fig. 6:** Comparative analysis of 3mm and 6mm dataset. (a)-(d) are BVC, VPI, BVT and BVD values for 3mm. (e)-(h) show BVC, VPI, BVT and BVD values for 6mm.

# Discussion

This study showcases the potential of AI to bridge the gap between OCT's inability to visualize blood flow information and leveraging generative-adversarial learning frameworks for image translation to capture that information. Our findings suggest that AI-driven translation models can generate high-quality OCTA images from OCT data (demonstrated using SSIM, FID and PCQI metrics) and the quantitative features generated in TR-OCTA follow a similar trend as in GT-OCTA. This has the potential to significantly improve the accuracy and efficiency of diagnosing and monitoring retinal diseases through OCTA imaging, emphasizing the need for further research and development in this area.

In this paper, we implement a recently demonstrated algorithm [36] for OCT-OCTA translation and validate the translated OCTA images to show their utility in quantitatively characterizing retinal features. We present a comprehensive analysis comparing the performance of GT- OCTA images with those generated by a TR-OCTA across different patient groups, including those with complete data sets, AMD, CNV, DR and normal cases for 2 datasets of 3mm and 6mm resolution. SSIM was utilized as a metric to assess the similarity between TR-OCTA and GT-OCTA images, providing insight into the translation model's ability to replicate key structural features of the retinal vasculature. For the complete dataset of 3mm (TABLE 1) comprising 200 patients, the mean SSIM for TR-OCTA was 0.4835, with a range of 0.29 to 0.60, indicating moderate similarity with GT-OCTA. However, for these TR images we considered two more quality metrics FID and PCQI scores which are more suitable for GAN generated image quality comparison against GT images. We found an FID score of 35.88 for 3mm dataset which is better in comparison to 6mm having a value of 49.06. On the other hand, PCQI scores showed a close similarity between both datasets with mean±st.d of 0.99795± 0.000457 and 0.99778± 0.000539 for 3mm and 6mm respectively.

BVD, BVC, BVT and VPI were evaluated, revealing slight discrepancies between TR-OCTA and GT-OCTA, with TR-OCTA exhibiting slightly higher BVD and lower VPI values. In the AMD subgroup (6 patients), TR-OCTA demonstrated a lower mean SSIM of 0.4513 (range: 0.29-0.55) compared to the complete dataset, suggesting a slight reduction in model performance in capturing the intricate vascular changes associated with AMD. This was further evidenced by the slight variations in BVD, BVC, BVT, and VPI between TR-OCTA and GT-OCTA, indicating the model's nuanced sensitivity to pathological alterations in retinal structures. For CNV patients (5 in total), the mean SSIM was 0.4754 (range: 0.44-0.52), reflecting a relatively better performance of the translation model in replicating the vasculature compared to the AMD group but still below the complete dataset's benchmark. This subgroup analysis underscores the model's potential in discerning and translating subtle vascular abnormalities characteristic of CNV. The DR subgroup (29 patients) showcased a mean SSIM of 0.4923 (range: 0.29-0.59), which is closer to the complete dataset's mean, suggesting that the model is relatively adept at mimicking DR-related vascular features. However, slight discrepancies in quantitative vascular metrics between TR-OCTA and GT-OCTA images were observed, indicating room for improvement in the model's accuracy. Lastly, the normal group (160 patients) displayed a mean SSIM of 0.4834 (range: 0.34-0.60), aligning closely with the complete dataset's mean SSIM. This similarity suggests that the translation model is quite effective in replicating normal retinal vasculature, as evidenced by the minor differences in vascular metrics compared to GT-OCTA.

In a detailed analysis of a 6mm OCTA dataset encompassing 300 patients, TR-OCTA images were evaluated for structural similarity against GT-OCTA images and same vascular metrics were analysed. The overall mean SSIM for the dataset was 0.4175, indicating a moderate resemblance to GT-OCTA images, with specific disparities in vascular metrics suggesting areas where the translation model could be improved. Subgroup analyses revealed nuanced differences in SSIM values across conditions like AMD, CNV, CSC, DR, RVO, other conditions and normal cases, with SSIM values ranging broadly from 0.3664 in RVO patients, highlighting significant translational challenges to 0.4329 in DR patients, where the model performed relatively better. These findings underscore the translation model's variable efficacy across different retinal conditions, pointing to the necessity for further refinement to more accurately capture and replicate the complex vascular features characteristic of various retinal diseases.

Overall, this analysis reveals that while the translation model holds promise in reproducing retinal vasculature across various conditions, there exist minor variations in the accuracy of vascular metrics between TR-OCTA and GT-OCTA images. These discrepancies underscore the necessity for ongoing enhancements to the translation model to achieve higher precision in vascular representation, particularly for pathological conditions where accurate vascular depiction is critical for clinical diagnosis and monitoring. One limitation of this study is that the number of patients varies widely from disease to disease therefore lacking generalization for different pathologies. Another limitation is the resolution of the vascular regions, which vary due to resolution of the dataset itself and dataset containing diseased as well as normal patients for both 3mm and 6mm.

In summary, this study demonstrates the potential of generative AI in enhancing OCT imaging for ophthalmic diagnostics. By validating quantitative features to check the viability of TR-OCTA, this research addresses significant limitations in widespread adoption of OCTA in clinical settings. Despite facing challenges such as generalization for different retinal diseases and difficulty in capturing detailed vascular networks, the study lays a solid foundation for future advancements in multi-modal OCT based retinal disease diagnosis and monitoring. The incorporation of AI not only promises to reduce the dependence on costly OCTA devices but also opens new avenues for accessible and accurate retinal healthcare solutions. Moving forward, it is imperative to refine these AI models to improve the resolution and accuracy of translated OCTA images, ensuring they can reliably support clinical decision-making and contribute to the broader understanding of retinal pathologies.

**Supplemental Materials:**

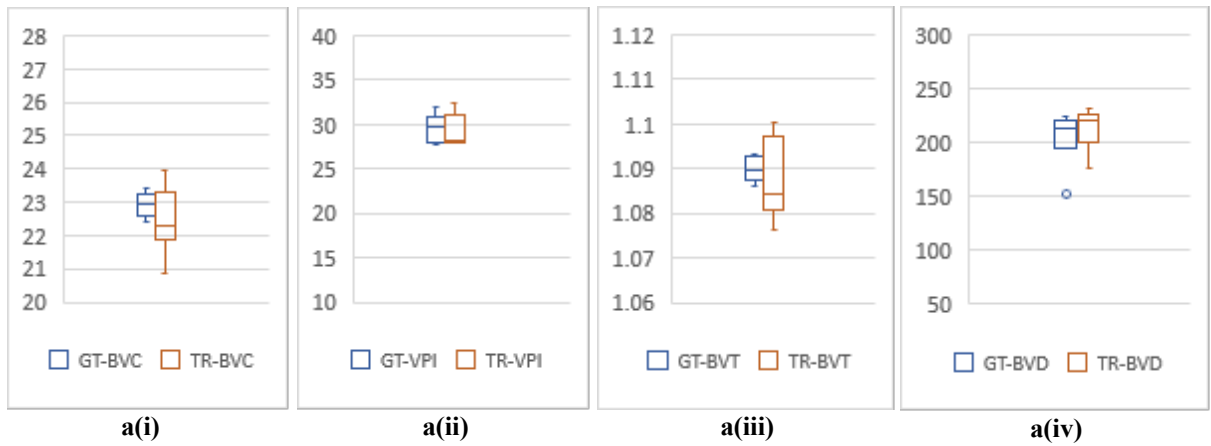

a(i)  a(ii)  a(iii)  a(iv)

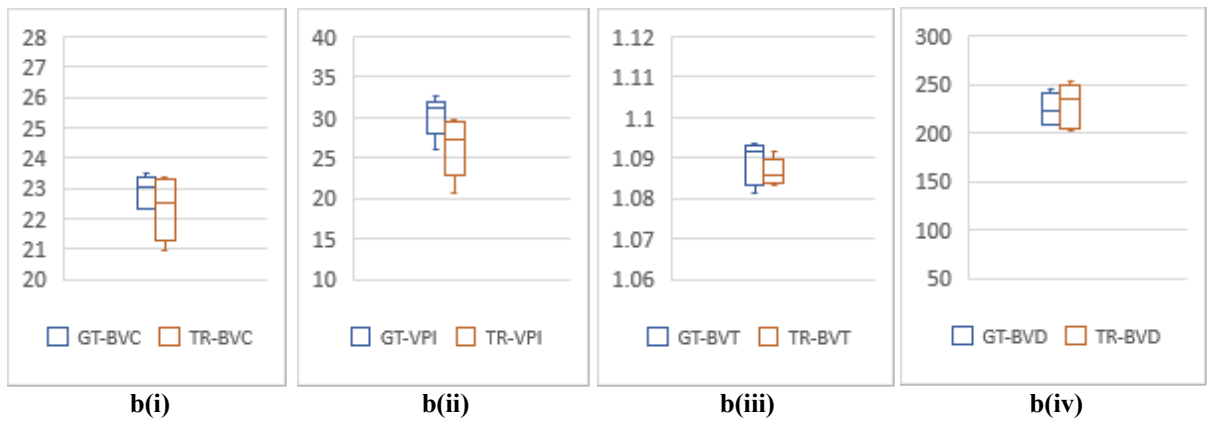

b(i)  b(ii)  b(iii)  b(iv)

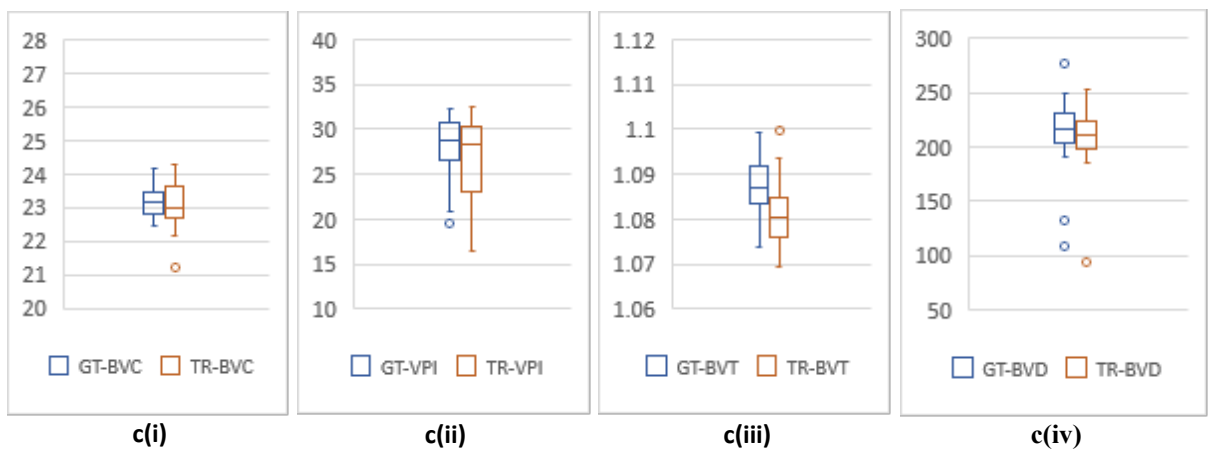

c(i)  c(ii)  c(iii)  c(iv)

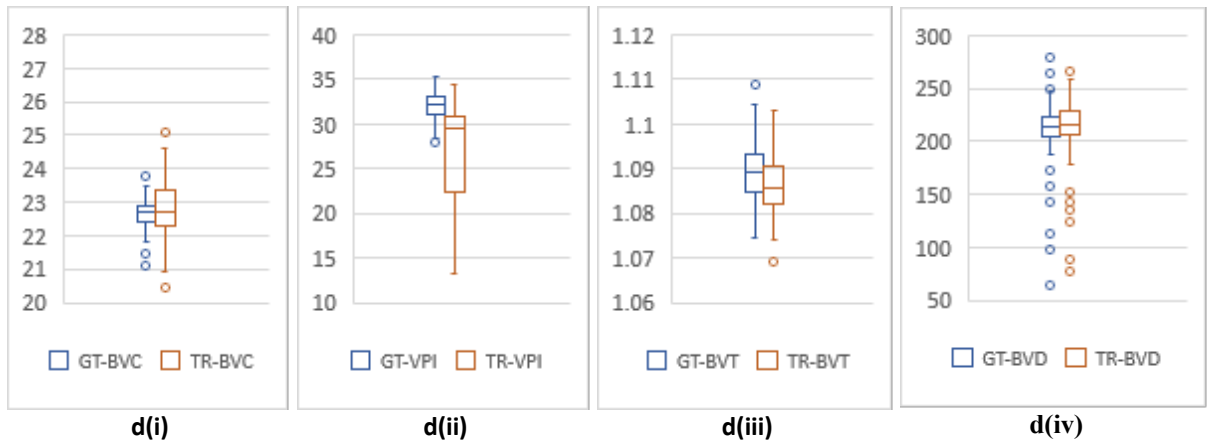

Supplemental Fig. 1: (a)-(d) show BVC, VPI, BVT and BVD for the 3mm dataset with different patient conditions. a(i-iv) are AMD patients, b(i-iv) are CNV patients, c(i-iv) are DR patients. d(i-iv) are Normal patients.

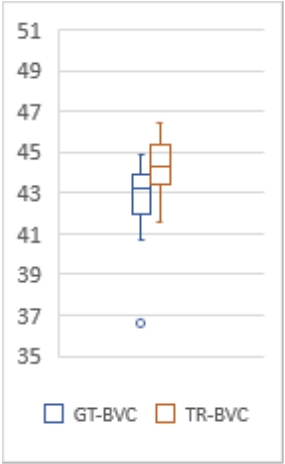 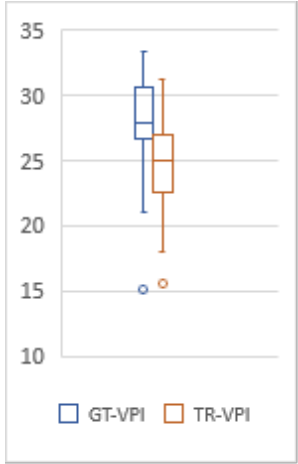 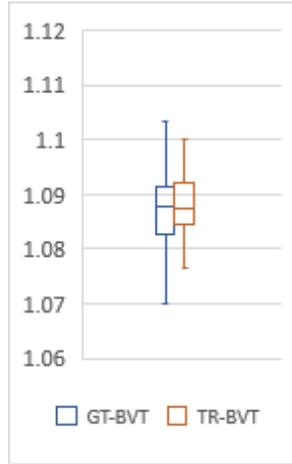 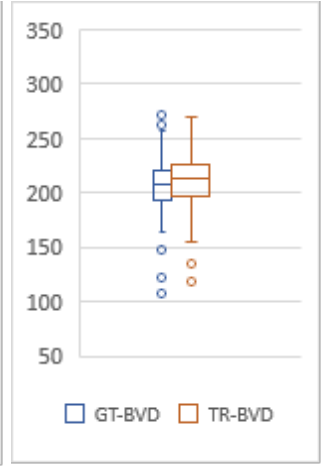

a(i)        a(ii)        a(iii)        a(iv)

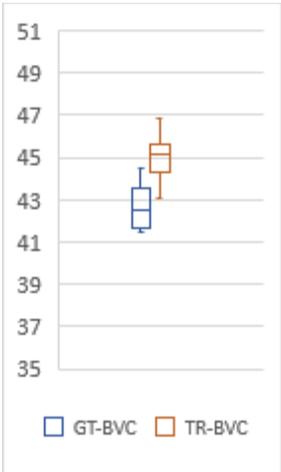 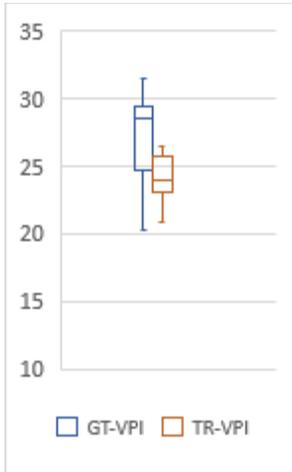 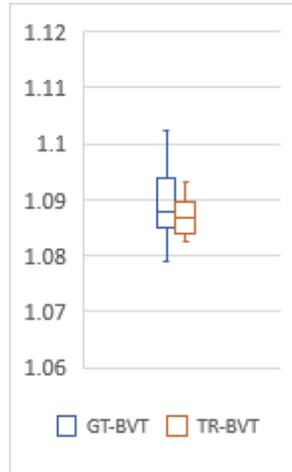 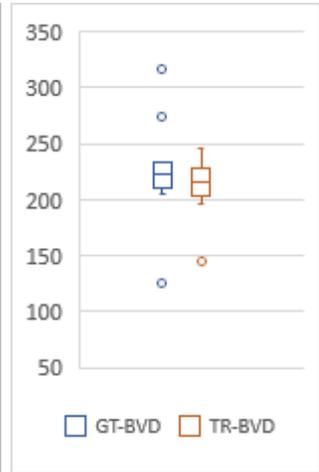

b(i)        b(ii)        b(iii)        b(iv)

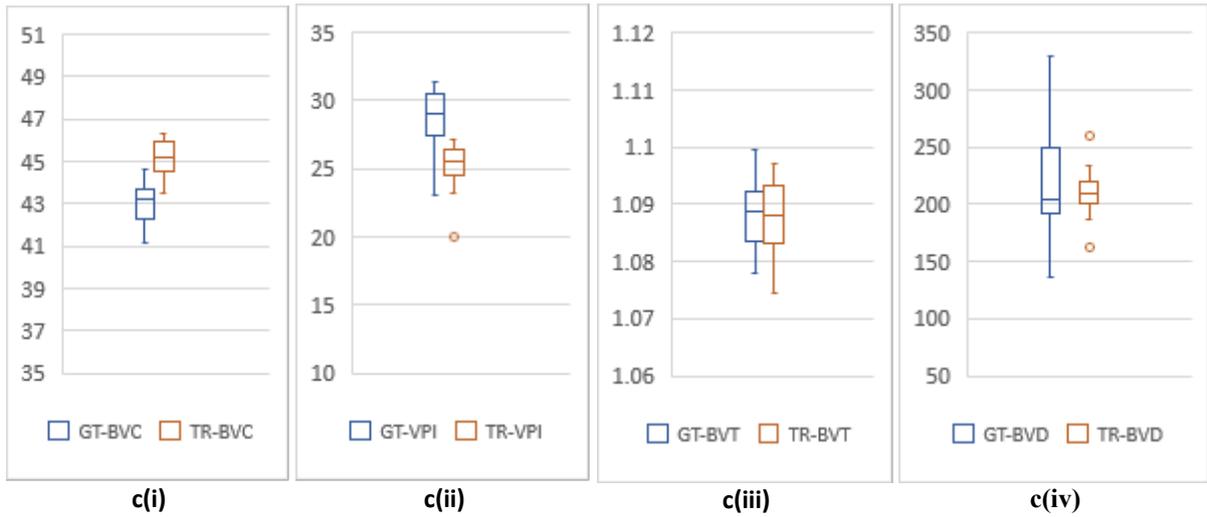
c(i) c(ii) c(iii) c(iv)

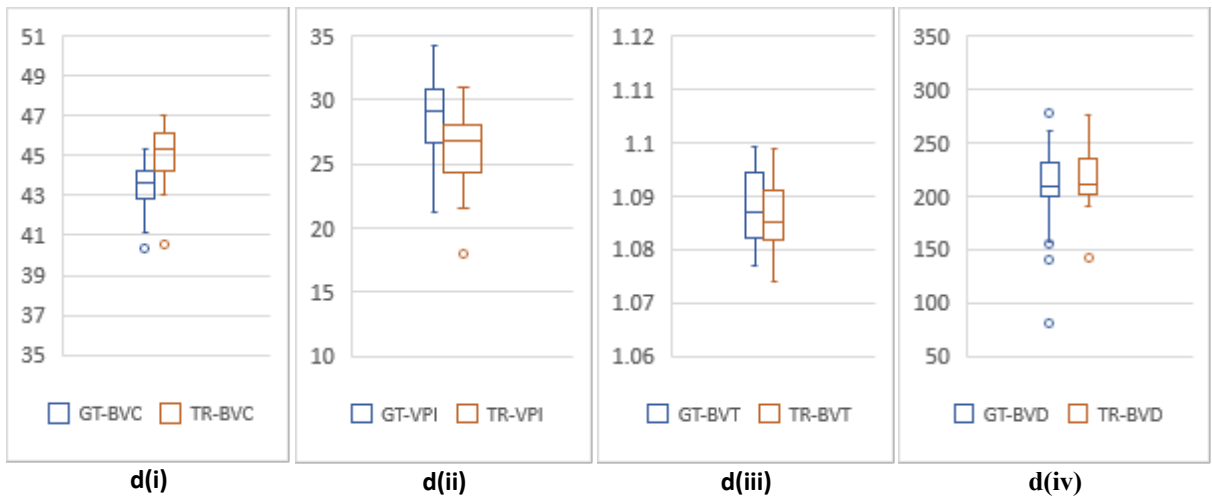
d(i) d(ii) d(iii) d(iv)

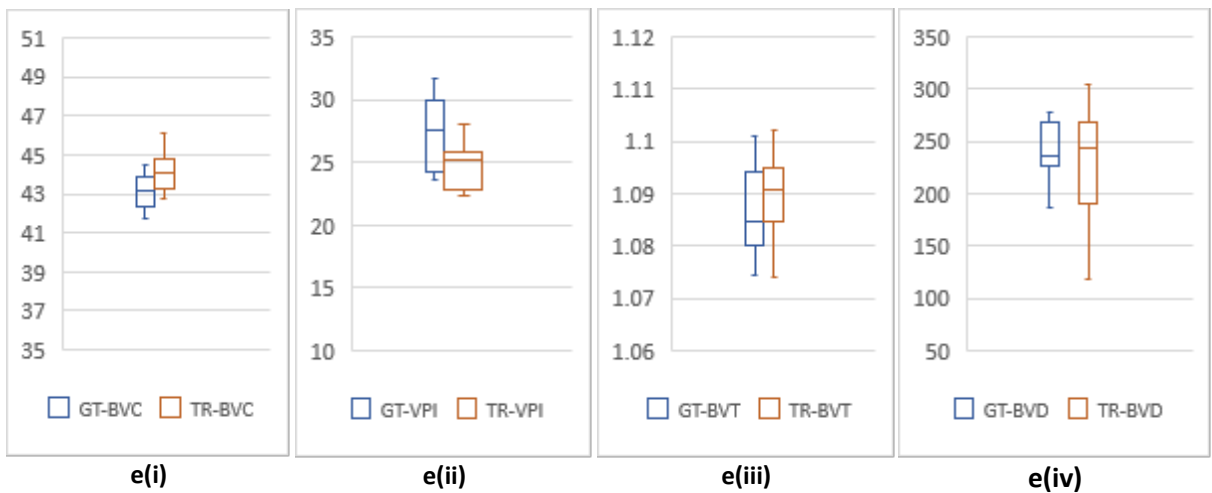
e(i) e(ii) e(iii) e(iv)

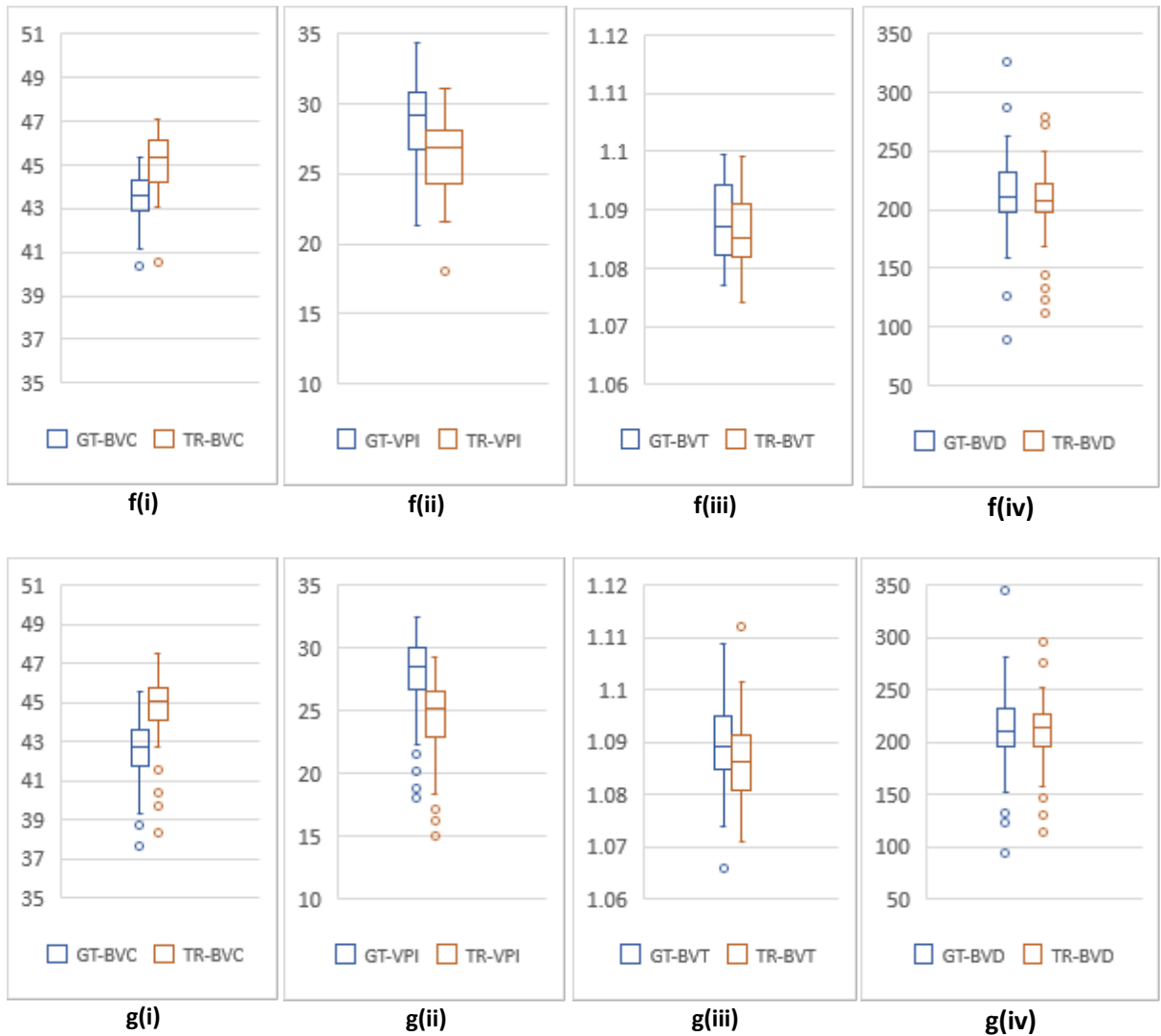

Supplemental Fig. 2: (a)-(g) show BVC, VPI, BVT and BVD for the 6mm dataset with different patient conditions. a(i-iv) are AMD patients, b(i-iv) are CNV patients, c(i-iv) are CSC patients, d(i-iv) are DR patients, e(i-iv) are RVO patients, f(i-iv) are patients with other retinal pathologies, g(i-iv) are Normal patients.